\documentclass{article} 
\usepackage{iclr2026_conference,times}

\usepackage{amsmath,amsfonts,bm}

\def\eqref#1{equation~\ref{#1}}
\def\Eqref#1{Equation~\ref{#1}}

\def\1{\bm{1}}

\DeclareMathAlphabet{\mathsfit}{\encodingdefault}{\sfdefault}{m}{sl}
\SetMathAlphabet{\mathsfit}{bold}{\encodingdefault}{\sfdefault}{bx}{n}

\usepackage{xcolor}
\usepackage{colortbl}
\usepackage{tablefootnote}
\usepackage{multirow}
\usepackage{booktabs,adjustbox,siunitx,array,threeparttable}
\usepackage{graphicx}
\usepackage{hyperref}
\usepackage{listings}
\usepackage{graphicx}
\usepackage{url}
\definecolor{scorercolor}{RGB}{218,194,177}
\definecolor{rlalgocolor}{RGB}{194,213,222}
\usepackage[most]{tcolorbox} 
\newenvironment{prompt}
  {\begin{tcolorbox}[colback=gray!4,colframe=gray!35,boxrule=0.4pt,
                     left=6pt,right=6pt,top=6pt,bottom=6pt]}
  {\end{tcolorbox}}
\usepackage[linesnumbered,ruled,vlined]{algorithm2e}
\title{SPIKE-RL: Video-LLMs meet Bayesian Surprise}
\author{%
  Sahithya Ravi $^{1,2}$ \hspace{2pt}
  Aditya Chinchure$^{1,2}$ \hspace{2pt}
  Raymond T. Ng$^{1}$\hspace{2pt}
  Leonid Sigal$^{1,2}$\hspace{2pt}
  Vered Shwartz$^{1,2}$ \\
   [1em]
    $^1$The University of British Columbia, \hspace{3pt} 
    $^2$Vector Institute for AI \\
\texttt{\{sahiravi, aditya10, rng, lsigal, vshwartz\}@cs.ubc.ca} 
}

\newcommand{\scorer}{\textsc{SPIKE}}
\newcommand{\rlalgo}{\textsc{SPIKE-RL}}

\newcommand{\code}[2]{%
  \begingroup\setlength{\fboxsep}{1pt}%
  \colorbox{#1}{#2}%
  \endgroup
}

\iclrfinalcopy 
\begin{document}

\maketitle
\begin{abstract}

Real-world videos often show routine activities punctuated by memorable, surprising events. However, most Video-LLMs process videos by sampling frames uniformly, likely missing critical moments that define a video's narrative. We introduce SPIKE, an inference-time framework that quantifies Bayesian Surprise as the belief update triggered by new visual evidence in the video stream, identifying moments where new visual evidence conflicts with prior beliefs. SPIKE effectively localizes surprise in videos, strongly correlated with humans on positive (FunQA) and negative (Oops!) surprise benchmarks. Since the beliefs of zero-shot Video-LLMs are often suboptimal, we develop SPIKE-RL, which leverages GRPO to optimize belief hypotheses based on a reward signal from the video caption. SPIKE and SPIKE-RL guide query-agnostic surprise-weighted frame sampling, which allocates more frames to interesting moments in the video. With this strategy, we achieve consistent performance gains on five downstream benchmarks over uniform sampling. By enabling Video-LLMs to track beliefs and register surprise, our work paves the way for more robust models that can revise their understanding in response to new information. Code is available at \url{https://github.com/sahithyaravi/SPIKE-RL}.
\end{abstract}

\section{Introduction}
\begin{figure}[t]
    \centering
    \includegraphics[width=1\linewidth]{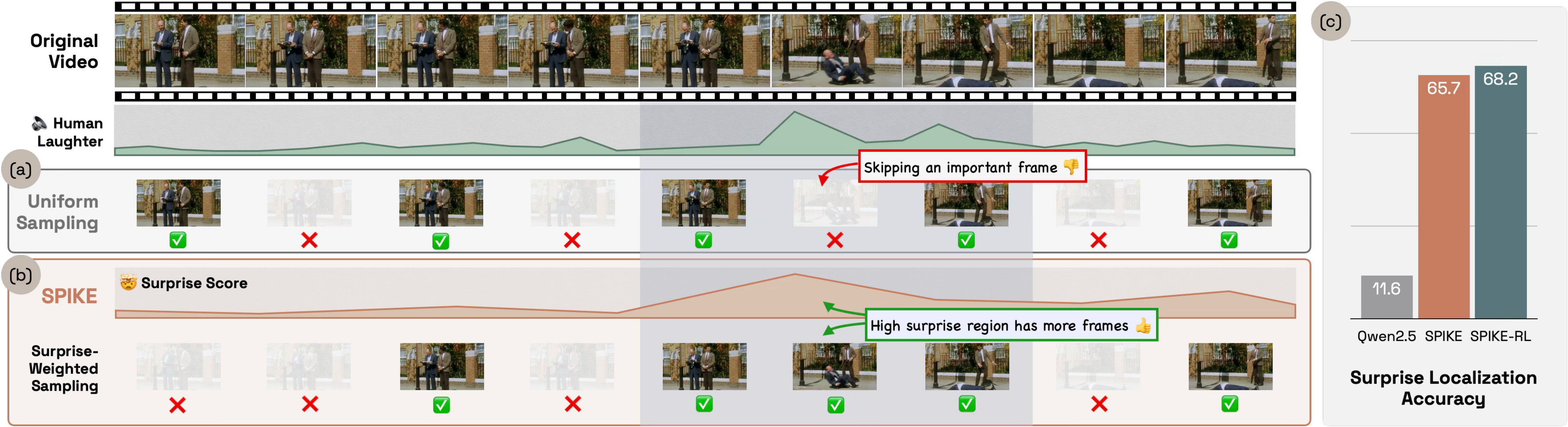}
    \caption{(a) Uniform sampling misses key moments. (b) Our surprise-based sampling focuses on high-surprise regions, strongly aligning with human laughter. (c) Our method achieves significantly better surprise localization than a zero-shot Qwen2.5-VL baseline.}
    \label{fig:teaser}
\end{figure}

Humans navigate the world not as passive observers, but as active predictors of the future  who infer the hidden causes behind events and update their predictions \citep{millidge2022predictivecodingtheoreticalexperimental}. This process, formalized within the Bayesian Theory of Mind (ToM) framework \citep{Baker2017RationalQA}, suggests that our brain continuously builds and updates an internal model of the world, using discrepancies between expectation and reality, or \textit{surprise}, as the primary signal for learning and attention. This allows us to efficiently process a constant stream of sensory data, focusing our cognitive resources on moments that are novel and informative, and ignoring redundant, expected information. For instance, in the Mr. Bean video shown in Figure~\ref{fig:teaser}, our cognitive focus is on the moment the man unexpectedly falls, because it deviates from the established routine. 

However, current Video-LLMs are fundamentally disconnected from this sequential, belief-driven process. Most models treat videos as a `bag of frames', where a subset is uniformly sampled from the video \citep{openai2024gpt4o, Qwen-VL, bai2025qwen25vltechnicalreport, cheng2024videollama2, liu2023improvedllava}. Lacking an evolving belief about the video's story, uniform sampling is much more likely to sample highly frequent mundane moments over rare surprising (and therefore memorable) events. This can potentially overwhelm Video-LLMs with redundant information, over pivotal moments a human observer would focus on, such as the fall in Figure ~\ref{fig:teaser}.

To overcome this, some methods select or retrieve frames retroactively for a given textual query \citep{yu2025framevoyager, Wang_2025_CVPR,Wang2024WeaklySG, Liang2024KeyVideoLLMTL, tang2025adaptive}. However, in dynamic, open‐world settings, we often don't know in advance what questions will be asked. What we need instead is a model that reasons \emph{proactively}, anticipating what is surprising, and paying attention to these shifts, similar to a human observer.  In this work, we study two fundamental questions to bridge this gap: (1) How can Video-LLMs proactively track and update their beliefs as new visual evidence presents itself? and (2) Can detecting semantically surprising events proactively and ahead of downstream queries improve video understanding? 


To answer these, we introduce \scorer, an inference-time framework that represents a model's beliefs as explicit probability distributions over human-interpretable textual hypotheses, and quantifies Bayesian Surprise as the divergence between prior and posterior beliefs \citep{NIPS2005_0172d289}, giving us a surprise score. As shown in Figure~\ref{fig:teaser}(b), this surprise score pinpoints moments that contradict the model's prior beliefs. We further improve the surprise scoring by introducing \rlalgo{}, trained using a reinforcement learning objective that teaches the model to prioritize beliefs that lead to more accurate video captions. \scorer\ achieves 65.7\% on FunQA \citep{xie2025funqa}, a surprise localization benchmark, and \rlalgo\ improves on it further, with 68.2\%, significantly outperforming the zero-shot performance of Qwen2.5-VL (Figure~\ref{fig:teaser}(c)).
Our experiments show that \rlalgo\ delivers two complementary benefits: it improves the diversity of generated belief hypotheses, and boosts surprise localization accuracy beyond what the inference-time scorer alone can achieve. Finally, we leverage this signal by replacing  the standard uniform frame sampling with surprise-weighted sampling in Qwen2.5-VL and demonstrate that this leads to consistent improvements on five downstream video understanding tasks. 


Our approaches allow Video-LLMs to focus on the most salient parts of the video, akin to human notions of surprise. In the future, surprise-aware Video-LLMs can be used to improve the robustness real-time applications such as streaming, surveillance, robotics, and interactive agents that need to adapt to new information on-the-fly.

\section{Bayesian Belief Tracking}
\label{sec:bayes-belief-tracking}

\subsection{Surprise Scoring}
\label{sec:sub:BeliefTracker}
\begin{figure}[t]
    \centering
    \includegraphics[width=\linewidth]{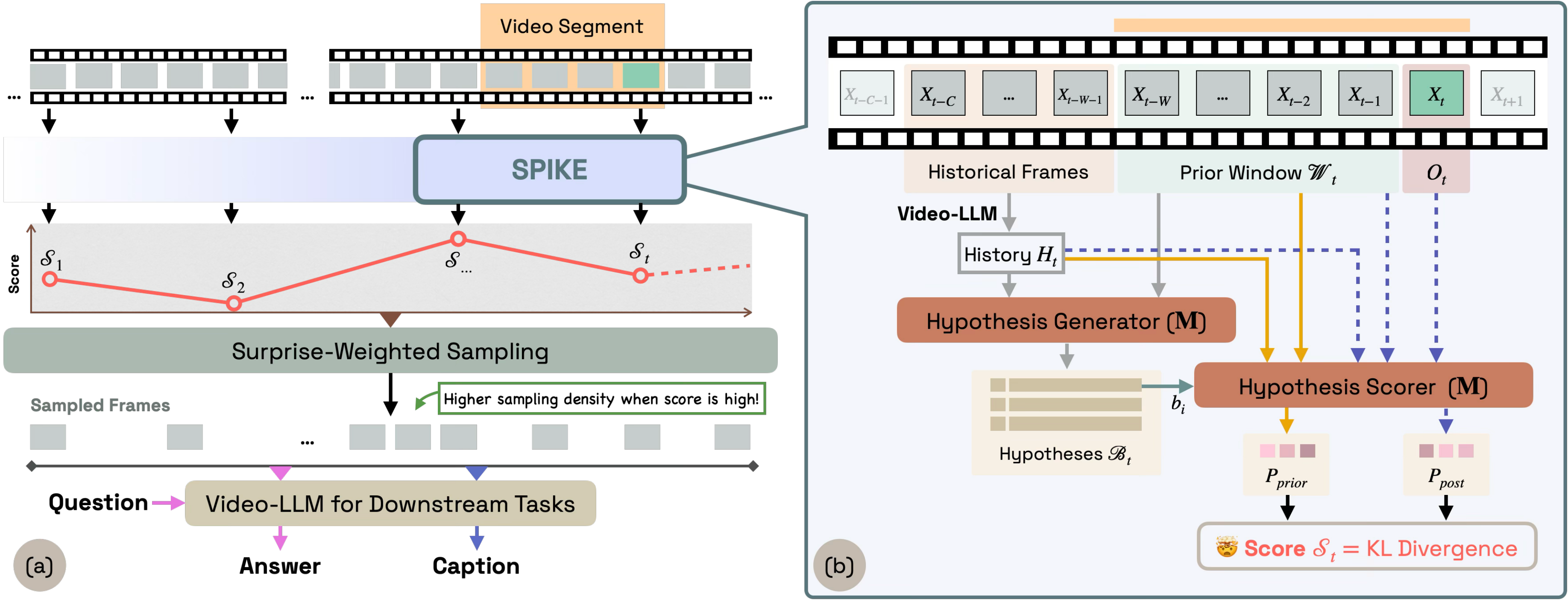}
    \caption{(a) Overall architecture: \scorer\ computes surprise scores, which guide weighted frame sampling for downstream tasks. (b) \scorer\ : Given history $H_t$, prior window $W$, and observed frame $O_t$, the hypothesis generator produces belief set $B_t$. The hypothesis scorer computes  $P_{prior}$ and  $P_{post}$, yielding surprise score $S_t$ as KL divergence.}
    \label{fig:main}
\end{figure}

The architecture of \scorer\ is shown in Figure \ref{fig:main}. \scorer\ quantifies Bayesian surprise by tracking how the model’s belief distribution over human-interpretable textual hypotheses shifts when a new frame is observed. Each incoming frame updates this belief distribution, and the magnitude of the change defines the surprise score. \scorer\ produces surprise scores for each step, across the complete video. For simplicity, we describe this process using fixed-length videos. However, our method can be adapted to a streaming video setup by applying the same update online.

\textbf{Setup.}
A video is composed of a sequence of frames $X_{1:T}$, where $T$ is the length of the video. To compute surprise at a timestep $t$, we use three key inputs as shown in Figure ~\ref{fig:main}(b): (i) the \emph{prior window} of $W$ frames immediately preceding the current $t$, $\mathcal{W}_t = X_{t-W:t-1}$, (ii) a \emph{historical summary}, $H_{t}$, a textual summary of what happened so far in the video, derived from the $C$ frames, $X_{t-C:t-W-1}$, that occurred before $\mathcal{W}_t$,\footnote{See Appendix \ref{app:historical_summary} for further information on how the textual summary is obtained.} and (iii) the newly observed frame $O_t=X_t$. This setup allows the model to form beliefs based on both long-term context and recent events, and then  measure surprise with respect to the new information.\footnote{See Appendix \ref{app:hypothesis_prompt} for the prompts for the hypothesis generation and scoring.}

\textbf{Hypothesis Generation.} First, at timestep $t$, we generate a set of belief hypotheses, $\mathcal{B}_t=\{b_{t,1},\dots,b_{t,N}\}$, where each hypothesis $b$ is a textual description of what might happen next, generated by a model $\mathbf{M}$ by conditioning on the historical summary $H_t$ and the prior frame window $\mathcal{W}_t$ (Fig. \ref{fig:main}). We use a Video-LLM as our model $\mathbf{M}$ and generate diverse beliefs $\mathcal{B}_t$ using nucleus sampling \citep{Holtzman2020The}.

\textbf{Bayesian Surprise.} Next, we establish \textbf{prior} and \textbf{posterior} belief distributions over the generated beliefs $\mathcal{B}_t$. We define a score for each hypothesis $b_{t,i}$ based on its plausibility, which is inversely proportional to its negative log-likelihood (NLL) as computed by the Video-LLM $\mathbf{M}$. This score reflects how well the hypothesis aligns with the given context. 

The prior distribution $P_{\text{prior}}$ is calculated based on the historical context ($H_t$) and the recent prior window ($\mathcal{W}_t$), \emph{before} the new frame $O_t$ is observed:
\begin{align}
    P_{\text{prior}}(b_{t,i} \mid H_t, \mathcal{W}_t) &= \frac{\exp\left( -\frac{1}{\tau} \cdot \text{NLL}(b_{t,i} \mid H_t, \mathcal{W}_t) \right)}{\sum_{j=1}^N \exp\left( -\frac{1}{\tau} \cdot \text{NLL}(b_{t,j} \mid H_t, \mathcal{W}_t) \right)},
\end{align}
where $\text{NLL}(b_i \mid \cdot) = -\log P_{\mathbf{M}}(b_i \mid \cdot)$ is the negative log-likelihood of the hypothesis tokens given the context, and $\tau$ is a temperature parameter. We apply softmax to normalize the scores into a probability distribution.

After observing the new frame $O_t$, we update our beliefs to form the posterior belief distribution, $P_{\text{post}}$, by incorporating this new visual evidence into the model's context:
\begin{align}
    P_{\text{post}}(b_{t,i} \mid H_t, \mathcal{W}_t, O_t) &= \frac{\exp\left( -\frac{1}{\tau} \cdot \text{NLL}(b_{t,i} \mid H_t, \mathcal{W}_t, O_t) \right)}{\sum_{j=1}^N \exp\left( -\frac{1}{\tau} \cdot \text{NLL}(b_{t,j} \mid H_t, \mathcal{W}_t, O_t) \right)}.
\end{align}

Following the Bayesian formalization of surprise by \citet{NIPS2005_0172d289}, we quantify our surprise score to be the information gain induced by $O_t$, as the Kullback–Leibler (KL) divergence between posterior and prior beliefs over hypotheses:
\begin{align}
\label{eq:bayesian-surprise}
\mathcal{S}_t \;&=\; D_{\mathrm{KL}}\!\left( P_{\text{post}}(\cdot \mid H_t,\mathcal{W}_t,O_t)\;\big\|\; P_{\text{prior}}(\cdot \mid H_t,\mathcal{W}_t)\right) \\
&= \sum_{i=1}^N P_{\text{post}}(b_{t,i})\,\log\frac{P_{\text{post}}(b_{t,i})}{P_{\text{prior}}(b_{t,i})}.
\end{align}

Using Equation \ref{eq:bayesian-surprise}, at each timestep $t$ we compute a scalar surprise score $\mathcal{S}_t$, as well as a belief set at $t$ containing hypotheses and their prior and posterior probabilities, $\mathcal{B}_t = \{(b_{t,i}, P_{\text{prior}}(b_{t,i}), P_{\text{post}}(b_{t,i}))_{i=1}^N\}_t$. $\mathcal{B}_t$ is human-readable and interpretable, enabling insight into \textit{why} a video segment is surprising. 




\subsection{Surprise-Weighted Frame Sampling}
\label{sec:sub:framealloc}
Since it is computationally infeasible and impractical to process all frames of a video, Video-LLMs sample frames -- by default, uniformly. Only the selected frames are then processed by the model while the rest are discarded. 
We define frame budget, $F$, as the maximum number of frames that a Video-LLM uses. Our goal is to effectively select those $F$ frames among the video frames $X_{1:T}$ by recognizing surprising regions of the video, which may be especially important for downstream tasks such as captioning and question answering.

\textbf{Computing a Surprise-Guided Probability Distribution.} As shown in Fig ~\ref{fig:main}(a), for a given video $X_{1:T}$, we first uniformly sample timesteps $t_1, \dots, t_K$, for $K\leq F$. Each timestep represents the end of a video segment, on which we measure surprise; this is akin to a sliding window over the frames of the video. We use \scorer\ to compute surprise scores for each segment, and obtain scores $\mathcal{S}_1,\dots,\mathcal{S}_K\in[0,1]$ for the corresponding timesteps $t_1, \dots, t_K$. We can now modify the frame sampling to be proportional to the surprise scores. Specifically, we compute the probability of sampling from a segment ending at $t_i$ as the softmax over scores, $
\label{eq:samplingPD}
p_i=\mathrm{softmax}\!\left(\frac{s_i}{\tau_s}\right)
=\frac{\exp(s_i/\tau_s)}{\sum_{j=1}^K \exp(s_j/\tau_s)} \quad(\tau>0),
$
and use $p_i{=}1/K$ if all $s_i$ are equal. $\tau_s$ is the temperature of this softmax function.

\textbf{Sampling.} Given the frame budget $F$ for the Video-LLM, we sample $F$ frames by repeatedly choosing a segment $i$ with probability $p_i$ (with replacement) and drawing a uniform timestamp within that segment; each timestamp is mapped to a frame index via the video frame rate. Choices are independent, so high-surprise segments can contribute multiple frames. We use $\tau_s$ in Eq. \ref{eq:samplingPD} to control sampling: a small $\tau_s$ concentrates the budget on surprising regions, whereas a larger $\tau_s$ spreads the frame budget more uniformly. We set $\tau_s = 0.7$ for our experiments.\footnote{App.~\ref{sec:complexity} provides the time complexity of the sampling approach.}

\section{Reinforcement Learning for Belief Optimization}
\label{sec:beliefRL}

\paragraph{Motivation.} The effectiveness of \scorer\ relies on the model's ability to generate belief hypotheses that are accurate, diverse, and representative of the video segment shown. However, since VLMs, are not tailored to perform belief tracking on frame windows,  the model has no incentive to refine its intermediate hypotheses. However, training \scorer\ with direct supervision on this reasoning process is intractable, as it is impractical to collect ground truth hypotheses across every segment of a video, for a large set of videos. Instead, we leverage GRPO \citep{shao2024deepseekmathpushinglimitsmathematical} to  optimize \scorer\ using reinforcement learning. \rlalgo{} is based on the insight that a strong final caption -- i.e. of what happened in the complete video -- is built upon accurate intermediate belief hypotheses -- i.e. about what is likely to happen after having watched a portion of the video. 

\begin{figure}[t]
    \centering
    \includegraphics[width=\linewidth]{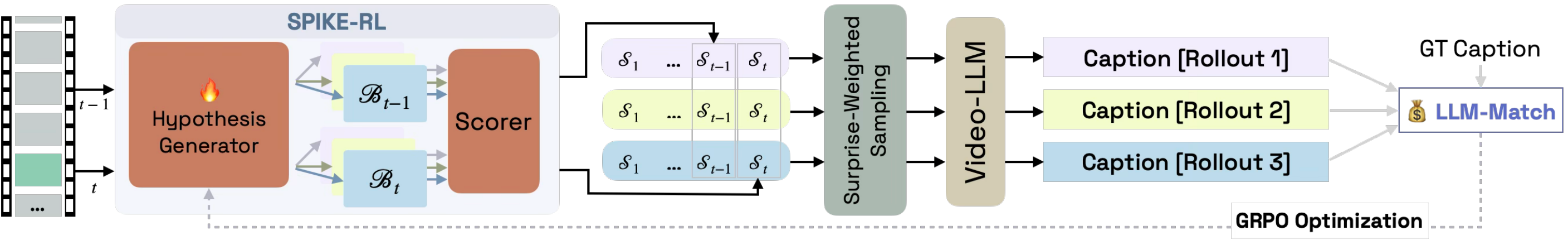}
    \caption{\rlalgo{} explores multiple hypothesis trajectories, whose surprise scores guide frame sampling. Captions from these rollouts are scored with LLM-Match, and GRPO propagates the reward to improve hypothesis generation. }
    \label{fig:grpo}
\end{figure}

Figure~\ref{fig:grpo} demonstrates our approach. To train the hypothesis generator, our policy model, we compute a reward signal based on the quality of the final caption. This reward signal is then propagated backward, assigning credit to the sequence of beliefs that led to the successful outcome. In this way, supervision on the final result is implicitly transformed into training feedback for the model's internal reasoning process. Our rewards are derived from an LLM-based metric that computes the similarity between the generated caption and the ground truth caption. 

\textbf{Rollout.} We design the GRPO-based training procedure by generating a group of captions, based on different \textit{trajectories} of beliefs and frame allocations. For each video, we draw $M$ \emph{trajectories} $\{\tau^{(r)}\}_{r=1}^M$. Each trajectory $\tau^{(r)}$ runs \scorer\ over segments of the video. At every timestep $t$, it samples $N$ textual beliefs $\mathcal{B}^{(r)}_t=\{b^{(r)}_{t,1},\dots,b^{(r)}_{t,N}\}$ and 
scores prior and posterior beliefs to obtain $\big(P^{(r)}_{\mathrm{prior},t}, P^{(r)}_{\mathrm{post},t}\big)$ and the surprise scores $\mathcal{S}^{(r)}_t$.
We then use the surprise scores to inform the sampling of frames that are inputted into a Video-LLM to generate a single final video caption, $c^{(r)}$ using our surprise-based frame allocation (\S \ref{sec:sub:framealloc}). Thus each input induces a GRPO \emph{group}: $
\mathcal{G} \;=\; \big\{\,(\{\mathcal{B}^{(r)}_t,\mathcal{S}^{(r)}_t\}_{t}^T,\; c^{(r)})\,\big\}_{r=1}^{M}$

\textbf{Reward.} At the end of a rollout, the caption $c^{(r)}$ is scored using LLM-Match, where an LLM judge measures how similar it is to the ground truth caption, to obtain a scalar reward $R^{(r)}$. The prompt for the LLM judge is in Appendix ~\ref{app:llm_reward}. We Z-score the LLM rewards within the group, and use the normalized scores as advantages in the policy objective,
$A^{(r)} = \frac{R^{(r)} - \mu_R}{\sigma_R}.$

\textbf{Loss.} We treat the full set of hypotheses in a trajectory as a sequence-level action. Let $p_\theta(b_{t,k}\mid H_t, \mathcal{W}_t)$ denote the policy for generating a hypothesis given the video context. We define our \textbf{belief-optimization} objective as,
\begin{equation}
\mathcal{L}_{\text{belief-optimization}}(\theta)
= -\frac{1}{M}\sum_{r=1}^M A^{(r)} 
\Bigg(\sum_t\sum_{k=1}^{K}\log p_\theta\!\big(b^{(r)}_{t,k}\,\big| H_t^{(r)}, \mathcal{W}^{(r)}_t \big)\Bigg),
\label{eq:belief-shape}
\end{equation}
which increases the likelihood of hypotheses along high-advantage trajectories and suppresses those along low-advantage ones. Optimizing \Eqref{eq:belief-shape} trains the model to generate hypotheses that reliably support strong captions, improving both the intermediate belief trajectory and the final output. 


\textbf{Training.} For training \rlalgo{}, we curated a video captioning dataset of 2,000 videos with 30\% \emph{surprising} and 70\% \emph{unsurprising} videos. The goal is to expose the policy both to routine events where beliefs remain stable and to inflection points that induce belief shifts. For the unsurprising portion, we used ActivityNet Captions \citep{caba2015activitynet}, which predominantly includes videos depicting everyday activities. For the surprising videos, we sample from from the training set of Oops! \citep{Epstein_2020_CVPR}, a collection of short clips centered on unintentional human failures. We use \texttt{Qwen2.5-VL-7B-Instruct} as the Video-LLM model ($\mathbf{M}$) and \texttt{Olmo-7B-hf} as the LLM-Match reward model. See App.~\ref{app:Hyperparams} for the training hyperparameters. 
\section{Surprise Localization}
\label{sec:Experiments}

\label{sec:task:local}

We first evaluate how well \scorer\ and \rlalgo\ can identify surprising segments of a video. Hyperparameters for surprise scoring are described in App.~\ref{app:Hyperparams}.

\subsection{Experimental Setup}

\textbf{Benchmarks.} We evaluate surprise localization on three benchmarks: Oops! \citep{Epstein_2020_CVPR}, FunQA \citep{xie2025funqa} and Mr.\ Bean (App. \ref{sec:app:bean}). Oops! is a surprise detection task, whose test set contains 4,791 videos with precise timestamps marking the exact transition point to surprise. FunQA has 424 videos with annotations for the most surprising segment in each video, given by a start and end time. While these are established benchmarks, they only annotate a single surprising event per video. Since our method is capable of detecting multiple surprising segments in the video, we curate our own benchmark, Mr.\ Bean, using 48 clips from the live-action TV show. Mr. Bean's audio laughter track serves as silver-standard surprise annotations -- segments of the video with laughter are considered surprising. 

\textbf{Metrics.} Following the protocols of Oops! and FunQA, we report Acc@0.25s
 and Acc@1.0s for Oops!, and IoU for FunQA. The accuracy metrics (Acc) measure whether the predicted surprise peak falls within 0.25 or 1.0 seconds of the ground truth peak surprise, while IoU measures the overlap between the predicted surprising windows and the ground-truth surprising windows. For details on the implementation of the metrics, see App. ~\ref{app:metrics_surprise}.

\textbf{Baselines.} We establish a lower bound with a Random baseline that selects surprising frames at random. We also report the zero-shot performance of our base \texttt{Qwen2.5-VL-7B-Instruct} model, which directly scores each uniformly sampled frame on whether it is surprising or not, without our proposed belief tracking mechanism (See  Appendix~\ref{app:qwen2.5_surprise_prompt} for the prompt and setup). On Oops!, we compare against: (i) VideoSpeed \citep{Epstein_2020_CVPR}, the strongest reported baseline for this dataset; (ii) Motion Magnitude \citep{Epstein_2020_CVPR}, an optical-flow-based approach; and (iii) F2C2V \citep{duka2022leveraging}, a self-supervised method. As an upper-bound reference, we also report the human consistency or agreement from the original dataset.
On FunQA, we compare against TimeChat \citep{Ren2023TimeChatAT}, UniVTG \citep{lin2023univtg}, a specialized video temporal grounding framework, and LLaVA-Next-CR, a baseline provided by the FunQA benchmark that applies the clipping and rating (CR) technique from UniVTG to LLaVA-NeXT \citep{liu2024llavanext}.

\subsection{Results} 
Table \ref{tab:results} shows the performance of \scorer\ and \rlalgo\ on the surprise localization task. On the Oops! benchmark, our \rlalgo\ model achieves an score of 62.9\% on Acc@0.25s, remarkably close to the human performance (62.1\%). Notably, both \scorer\ and \rlalgo\ show about a tenfold improvement over the performance of the zero-shot version of the same model (\texttt{Qwen2.5-VL-7B}). Compared to VideoSpeed, F2C2V, we observe that \scorer\ and \rlalgo\ are better at accurate localization, with a 23.4\% higher Acc@0.25s, and achieve similar Acc@1s scores. On the FunQA benchmark, \rlalgo\ once again demonstrates superior performance with an IoU of 68.2, surpassing both prior approaches and the zero-shot model by a substantial margin. It is worth noting that this significant boost is despite the fact that FunQA -- which is composed of positive surprises related to humor and creativity -- is out-of-distribution for \rlalgo{}.  


Mr.\ Bean shows a similar trend to the other benchmarks, but the absolute Acc@0.25s is lower. This dataset is particularly challenging. In contrast to the other benchmarks, some of the surprising moments in Mr.\ Bean arise from subtle, fine-grained nuances in his facial expressions rather than easily noticeable unexpected events. Finally, we observe a significant 6.3\% gain in IoU score with \rlalgo\ over \scorer. Since IoU on Mr.\ Bean evaluates detection across multiple surprising segments, this gain highlights the ability of our scorer to capture nuanced surprises within a video.

Overall, the inference-time method, \scorer, achieves superior performance across all benchmarks and generalizes to different types of surprises, while \rlalgo\ further boosts performance through reinforcement-guided refinement.

\begin{table}[t]
\centering
\caption{Performance of \scorer\ and \rlalgo\ on surprise localization. }
\label{tab:results}
\setlength{\tabcolsep}{6pt}
\renewcommand{\arraystretch}{1}
\begin{adjustbox}{width=0.8\textwidth,center}
\begin{tabular}{
  >{\raggedright\arraybackslash}p{0.19\linewidth}
  *{2}{c@{\hspace{8pt}}} 
  *{1}{c@{\hspace{8pt}}} 
  *{3}{c}                
}
\toprule
& \multicolumn{2}{c}{\textbf{Oops!}} 
& \multicolumn{1}{c}{\textbf{FunQA}} 
& \multicolumn{3}{c}{\textbf{Mr.\ Bean}} \\
\cmidrule(lr){2-3}\cmidrule(lr){4-4}\cmidrule(lr){5-7}
\textbf{Method} & 
\textbf{Acc@0.25s} & \textbf{Acc@1s} &
\textbf{IoU} &
\textbf{Acc@0.25s} & \textbf{Acc@1s} & \textbf{IoU} \\
\midrule
\multicolumn{7}{l}{\emph{Baselines}} \\

\addlinespace[2pt]
Random & 6.8 & 2.6 & 7.5 & 0.6 & 3.5 & 0.9 \\
Motion & 23.1 & 50.7 & -- & -- & -- & -- \\
Video Speed & 36.6 & 65.3 & -- & -- & -- & -- \\
F2C2V  & 39.5 & \textbf{69.5} & -- & -- & -- & -- \\
TimeChat & -- & -- & 9.6 & -- & -- & -- \\
UniVTG & -- & -- & 45.3 & -- & -- & -- \\
LLaVA-NeXT-CR & -- & -- & 62.3 & -- & -- & -- \\
Qwen2.5-VL & 6.6 & 9.6 & 11.6 & 11.2 & 23.2 & 13.8 \\
\midrule
\addlinespace[4pt]
\rowcolor{scorercolor!30} \scorer & 60.0 & 67.3 & 65.7 & 53.2 & 70.2 & 54.8 \\
\rowcolor{rlalgocolor!30} \rlalgo & \textbf{62.9} & 69.1 & \textbf{68.2} & \textbf{57.4} & \textbf{78.7} & \textbf{61.1} \\
\midrule
\addlinespace[2pt]
Human & 62.1 & 88.0 & -- & -- & -- & -- \\
\bottomrule
\end{tabular}
\end{adjustbox}
\end{table}

\subsection{Belief Set Evaluation}
We evaluate the hypotheses generated by \scorer\ and \rlalgo\ using a combination of automatic metrics and human evaluation. 

\textbf{Diversity.} 
We are interested in whether models generate multiple conceptually-diverse hypotheses or different lexical variations of the same hypothesis. For a given video, we measure the diversity of a hypothesis set with the average inverse cosine similarity ($1-cos(b
_i, b_j)$) across all hypothesis pairs. 
\rlalgo\ achieves \code{rlalgocolor!30}{40.3\%}, higher than \scorer’s \code{scorercolor!30}{33.5\%}, showing that the RL training improves diversity.

\textbf{Correlation with human judgments.} 
We measure how well our surprise score aligns with human judgments by showing human annotators a random sample of  100 videos from Oops! along with the generated hypotheses and asking them to rank the hypotheses by dragging them onto a 0--100 scale. Each video segment is evaluated twice: first using only the prior frames ($O_{<t}$), and then again after revealing the observed frame ($O_t$). This setup emulates the prior and posterior probabilities in Eq.~\ref{eq:bayesian-surprise}, from which we compute human-derived surprise scores. Comparing these to \scorer\ and \rlalgo's surprise scores  yields a Spearman correlation of \code{scorercolor!30}{0.84} and \code{rlalgocolor!30}{0.87} respectively, indicating \textbf{very strong correlation} and confirming that our method effectively captures belief shifts. The template for human evaluation is provided in App.~\ref{app:human_eval_temp}.


\begin{table}[t]
\centering
\caption{Performance of Qwen2.5-VL with uniform vs. surprise-weighted and other query-free frame sampling methods.
MCQ tasks are evaluated with accuracy; generative tasks with LLM-Match. Comparable open-source Video-LLMs are shown for context.}
\label{tab:vlm_comparison}
\setlength{\tabcolsep}{7pt}
\renewcommand{\arraystretch}{1.1}
\begin{adjustbox}{width=\textwidth,center}
\begin{tabular}{
  >{\raggedright\arraybackslash}p{0.18\linewidth}
  c
  >{\raggedright\arraybackslash}p{0.18\linewidth}
  c@{\hspace{6pt}}c@{\hspace{6pt}}c
  |
  c@{\hspace{6pt}}c
}
\toprule
\textbf{Model} & \textbf{Size} & \textbf{Sampling} & 
\textbf{BlackSwan} & \textbf{FunQA} & \textbf{ExFunTube} & 
\textbf{VideoMME-S} & \textbf{NextQA} \\
\midrule
\addlinespace[3pt]
VideoChat2 & 7B & Uniform &  49.7 & 17.9 & -- & 45.6 &  -- \\
VideoLlama2 & 7B & Uniform & 52.9 & 7.7 & -- & 56.0 &  --\\
FunMentor & 7B & Uniform & -- & 33.2 & -- & -- & --  \\
LLaVA-Video & 7B & Uniform & 70.4 & -- & -- & 46.6 & 62.7 \\
\midrule
Qwen2.5-VL & 7B & Uniform & 67.2 & 66.8 & 68.7 & 59.8 & 68.6\\
Qwen2.5-VL &  7B & RGB Histogram & 49.6 & --& -- & 55.4 & --\\
Qwen2.5-VL &  7B & ECR & 49.7 &-- & -- & 58.2 & --\\
Qwen2.5-VL &  7B & Katna & 54.6 & -- &--  & 57.4 & --\\
Qwen2.5-VL &  7B & Optical Flow & 58.6 &-- & -- & 58.1 &-- \\
\rowcolor{scorercolor!30}  Qwen2.5-VL & 7B & \scorer & 68.8 & 70.3 & 73.2 & 60.8 & 69.8 \\
\rowcolor{rlalgocolor!30} Qwen2.5-VL & 7B & \rlalgo & 69.5 & 71.4 & 75.7 & 62.5 & 70.3\\
\midrule
Qwen2.5-VL & 32B & Uniform & 69.4 & 72.7 & 71.9 & 69.9 & 72.3\\
\rowcolor{rlalgocolor!30} Qwen2.5-VL & 32B & \rlalgo & \textbf{71.7} & \textbf{75.8} & \textbf{75.8} &  \textbf{73.5} & \textbf{74.1}\\

\bottomrule
\end{tabular}
\end{adjustbox}
\end{table}

\section{Downstream Tasks}
\label{sec:task:qa}

Having shown that \scorer\ and \rlalgo\ can perform surprise localization, we now explore how identifying surprising segments of the video and allocating more frames to such regions can improve a Video-LLM's performance on downstream tasks as described in \S\ref{sec:sub:framealloc}. 

\begin{figure}[t]
    \centering
    \includegraphics[width=1\linewidth]{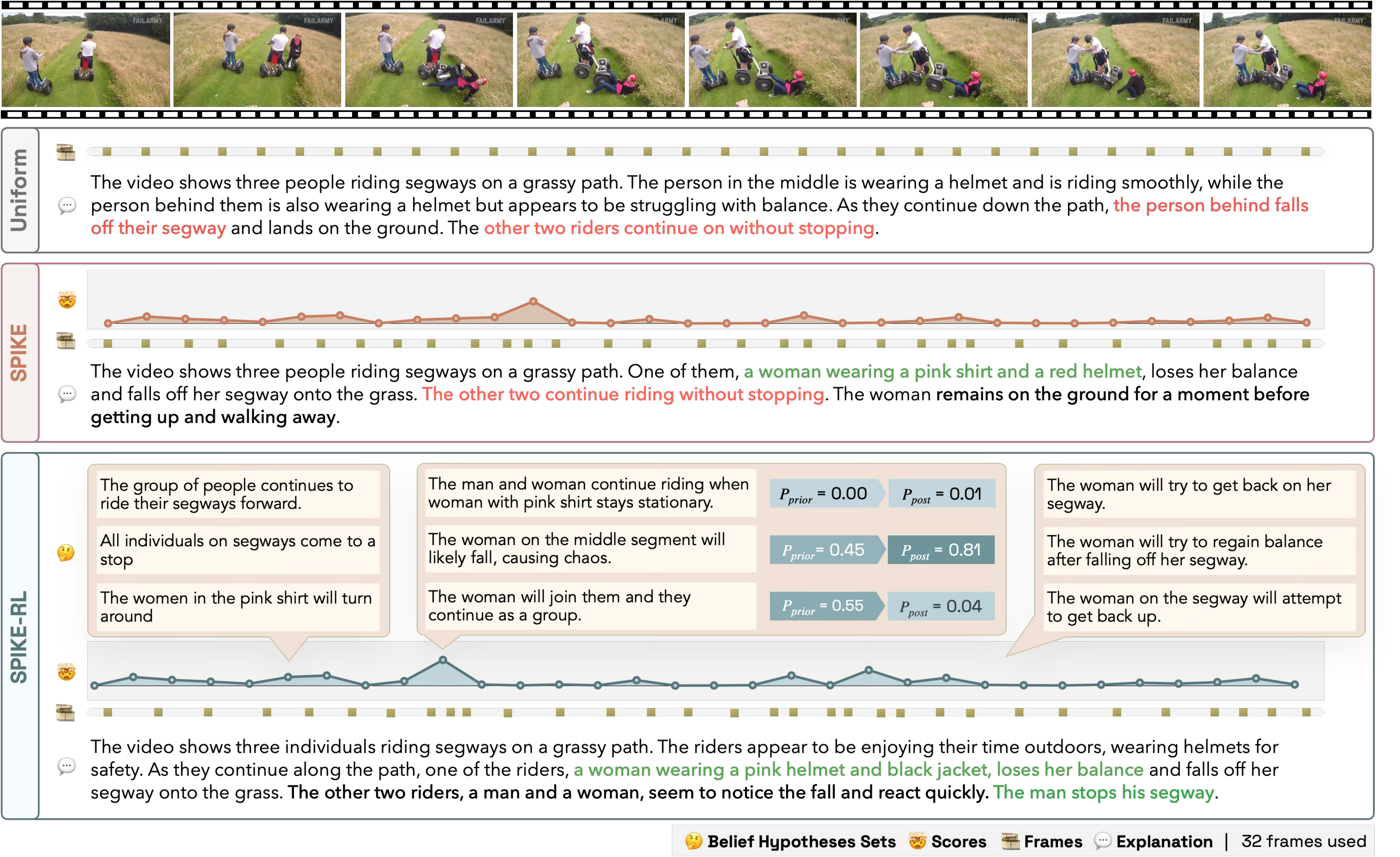}
    \caption{\textbf{Qualitative Results.} We compare Uniform, \scorer\ and \rlalgo\ sampling methods. Errors in the explanation generated using uniform sampling reduce with \scorer\ and are resolved with \rlalgo. We show belief hypotheses sets ($\mathcal{B}_t$) at various timesteps, and observe how the divergence of $P_{prior}$ and $P_{post}$ accurately captures the surprising moment in the video.
}
    \label{fig:qualitative}
\end{figure}

\subsection{Experimental Setup}
\textbf{Benchmarks.} We evaluate our sampling method on a diverse selection of tasks, spanning surprise explanations, question answering, and temporal reasoning. The Reporter-MCQ portion of BlackSwanSuite \citep{chinchure2025black} tests models' ability to describe an unexpected event in a MCQ setup. FunQA's Task 2 \citep{xie2025funqa} and ExFunTube \citep{ko2023can} ask models to generate an explanation of why events are surprising. 
Moving beyond surprising videos, we test our models on two MCQ tasks -- VideoMME \citep{fu2024video}, which probes general multimodal reasoning (we focus on short videos without subtitles), and NextQA \citep{xiao2021next}, which targets temporal, commonsense, and causal reasoning. 

\textbf{Metrics.} Following prior work \citep{majumdar2024openeqa,xie2025funqa}, we evaluate the generative tasks using LLM-Match, prompting GPT-4o to rate the similarity between model-generated and ground-truth answers. Multiple-choice tasks are evaluated using accuracy. 

\textbf{Video-LLM Baselines}. We consider widely adopted open-source Video-LLMs capable of video explanation and QA, including VideoChat2 \citep{li2024mvbenchvideochat2}, VideoLlama2 \citep{cheng2024videollama2}, and LLaVA-Video \citep{liu2023improvedllava}. We also include FunMentor \citep{xie2025funqa}, a model specifically designed for humor understanding. Our base model is Qwen2.5-VL (7B), which we use to evaluate alternative sampling strategies under a fixed frame budget on BlackSwan and VideoMME-S. Finally, we test whether \rlalgo\ improves performance on a larger model, Qwen2.5-VL (32B).

\textbf{Query-free Frame Sampling Baselines}. To assess the effectiveness of our sampling, we benchmark against shot boundary detection methods on BlackSwan and Video-MME-S. Specifically, we tested RGB Histogram differences \citep{CV2015KeyframeEB}, Edge Change Ratio  \citep[ECR;][]{Mann2015KeyFE}, and motion-based detection \citep{Wolf1996KeyFS}, which capture changes in texture, structure, motion, and similarity. In all of these approaches, salient peaks are detected via smoothed scores and frames are distributed proportionally to peak strength, ensuring that the frame budget $F$ is met. We also include Katna,\footnote{\url{https://github.com/keplerlab/katna}} a clustering-based method  which applies K-means to frame histograms and selects the frame closest to each centroid. We use a maximum frame budget $F$ of 64 frames for all our baselines, regardless of the sampling approach.


\subsection{Results} 
Table~\ref{tab:vlm_comparison} shows the performance of \scorer\ and \rlalgo\ on downstream benchmarks. On tasks with surprising videos (BlackSwan, FunQA, ExFunTube), surprise-aware sampling provides substantial gains over uniform selection. Relative to uniform sampling, \scorer{} improves accuracy by +1.6\% on BlackSwan, +3.5\% on FunQA, and +4.5\% on ExFunTube. We observe that \rlalgo\ further extends performance on these tasks, with gains of +2.3\% and +4.6\% on BlackSwan and FunQA, and +7.0\% on ExFunTube, marking our largest gains over uniform sampling. These results not only show the effectiveness of \scorer\ in prioritizing surprising frames, but also credit the improved hypothesis quality in \rlalgo. 
On Qwen2.5-VL 32B, we see 2.3\%, 3.1\% and 3.9\% gains respectively with \rlalgo, showing that our methods benefit larger models as well, extending their video understanding capability.

In general QA tasks (VideoMME-S, NextQA), we see moderate but consistent improvements over uniform sampling. \scorer{} boosts scores by +1.0\% on VideoMME-S and +1.2\% on NextQA, while \rlalgo{} achieves +2.7\% and +1.7\% respectively on the 7B variant. The 32B variant with \rlalgo\ shows larger improvements of 3.6\% and 1.8\% on these tasks. These results show that surprise-aware sampling is broadly beneficial. 

SBD strategies such as RGB Histogram, ECR, Katna, and Optical Flow consistently underperform uniform sampling. Their reliance on raw visual change makes them sensitive to camera motion and scene cuts, which rarely align with semantically important events. In contrast, our method offers principled guidance for identifying critical moments. Overall, we demonstrate that Bayesian Surprise provides a powerful inductive signal for adaptive frame selection: \scorer\ delivers immediate gains by reallocating a fixed frame budget toward more informative segments, while \rlalgo{} further improves robustness through reinforcement-guided belief optimization. 




\subsection{Qualitative Example} 

Figure~\ref{fig:qualitative} illustrates the differences between uniform sampling, \scorer{}, and \rlalgo. Under uniform sampling, the Video-LLM generates a caption that notes someone falling off a segway but misidentifies the person and the actions of the other riders (error highlighted in red). With the same frame budget, \scorer\ and \rlalgo\ reallocate samples toward segments with high surprise scores, guided by observed belief shifts as demonsrated by the hypotheses. \scorer\ correctly captures that the woman in the pink shirt and helmet loses balance and falls, though it still makes an error by stating that the other riders continue without stopping. \rlalgo{} improves on this. By more accurately localizing surprising segments -- with one peak at the main fall and another smaller peak later -- \rlalgo{} increases sampling density around both critical events. This leads to a more precise description of both the fall and the subsequent reactions of the other riders.

\section{Related Work}
\label{sec:related_work}

\textbf{Belief Tracking and Updating.} 
Recent research in NLP has explored the idea of maintaining and updating beliefs, often with Bayesian inspired methods. Studies show that, with sufficient evidence, LLMs can approximate Bayesian updates rather than simply relying on pattern matching \citep{gupta-etal-2025-enough}. Closest to our work, \cite{kim2025hypothesisdriven} explicitly maintain and re-weight hypotheses about agents' mental states as new information becomes available, mirroring Bayesian Theory of Mind.  This principle of explicit tracking also improves model robustness in complex scenarios involving multiple characters and higher-order Theory of Mind \citep{sclar-etal-2023-minding}. This process is closely related to the concept of defeasible reasoning, where conclusions are revised by new evidence \citep{rudinger-etal-2020-thinking}. More broadly, the principle of Bayesian Surprise has been used as a powerful driver for exploration in other domains, such as guiding open-ended scientific discovery \citep{agarwal2025openendedscientificdiscoverybayesian}. We extend this idea of discovery to the domain of video understanding. 

\textbf{Adaptive Frame Sampling Strategies for Video-LLMs.}
Prior work on frame selection for Video-LLMs is primarily based on the relevance to the question. Query-conditioned methods rank frames with respect to a textual prompt to minimize redundancy while preserving task-relevant evidence. Frame-Voyager\citep{yu2025framevoyager}, Flexible Frame Selection \citep[FFS;][]{Buch2025FlexibleFS} and \cite{Hu2025MLLMBV} learn to select informative frame sets conditioned on the query using lightweight training strategies. Adaptive Keyframe Sampling \citep[AKS;][]{tang2025adaptive} formulates selection as a plug-and-play module optimizing relevance to the prompt and \cite{guo2025logicinframesdynamickeyframesearch} propose dynamic keyframe search driven by visual chain-of-thought. VideoTree \citep{Wang_2025_CVPR} organize a video into a hierarchical tree and traverse it in a question adaptive manner. In contrast, our method drops in as a replacement for the Video-LLM’s uniform sampling layer, reallocating the frame budget towards surprising moments while remaining query-agnostic. 

\section{Conclusion}
\label{sec:conclusion}
We introduced \scorer, a framework that lets Video-LLMs proactively register surprise. We further showed that \rlalgo\ can refine intermediate belief generation, improving both belief diversity and surprise localization. This enables surprise-driven frame sampling, yielding consistent gains across downstream tasks, especially when critical information is sparse. Modeling surprise offers a path toward understanding video narratives, adapting to violated expectations, and anticipating events. Future work could investigate extending \scorer\ to real-time streams and combining with task-specific relevance signals.
\section{Reproducibility Statement}

We intend to make all our data, code and models open-source. \scorer\ is based on an open source Video-LLM, Qwen2.5-VL, and our training code for \rlalgo\ will be made available on GitHub. We note that all our prompts are included in Appendix \ref{app:allprompts} and hyperparameters in Appendix \ref{app:Hyperparams}. For the Mr.\ Bean evaluation set we create, we plan to share the video clips, along with annotations containing their original source. We also share the instructions and template used in our human evaluation in Appendix \ref{app:human_eval_temp}.

\section*{Acknowledgement}
This work was funded, in part, by the Vector Institute for AI, Canada CIFAR AI Chair, NSERC CRC, and NSERC DG. Hardware resources used in preparing this research were provided, in part, by the Province of Ontario, the Government of Canada through CIFAR, and companies sponsoring the Vector Institute.
\bibliography{iclr2026_conference}

\begin{thebibliography}{41}
\providecommand{\natexlab}[1]{#1}
\providecommand{\url}[1]{\texttt{#1}}
\expandafter\ifx\csname urlstyle\endcsname\relax
  \providecommand{\doi}[1]{doi: #1}\else
  \providecommand{\doi}{doi: \begingroup \urlstyle{rm}\Url}\fi

\bibitem[Agarwal et~al.(2025)Agarwal, Majumder, Adamson, Chakravorty, Gavireddy, Parashar, Surana, Mishra, McCallum, Sabharwal, and Clark]{agarwal2025openendedscientificdiscoverybayesian}
Dhruv Agarwal, Bodhisattwa~Prasad Majumder, Reece Adamson, Megha Chakravorty, Satvika~Reddy Gavireddy, Aditya Parashar, Harshit Surana, Bhavana~Dalvi Mishra, Andrew McCallum, Ashish Sabharwal, and Peter Clark.
\newblock Open-ended scientific discovery via bayesian surprise, 2025.
\newblock URL \url{https://arxiv.org/abs/2507.00310}.

\bibitem[Bai et~al.(2023)Bai, Bai, Yang, Wang, Tan, Wang, Lin, Zhou, and Zhou]{Qwen-VL}
Jinze Bai, Shuai Bai, Shusheng Yang, Shijie Wang, Sinan Tan, Peng Wang, Junyang Lin, Chang Zhou, and Jingren Zhou.
\newblock Qwen-vl: A frontier large vision-language model with versatile abilities.
\newblock \emph{ArXiv preprint}, 2023.

\bibitem[Bai et~al.(2025)Bai, Chen, Liu, Wang, Ge, Song, Dang, Wang, Wang, Tang, Zhong, Zhu, Yang, Li, Wan, Wang, Ding, Fu, Xu, Ye, Zhang, Xie, Cheng, Zhang, Yang, Xu, and Lin]{bai2025qwen25vltechnicalreport}
Shuai Bai, Keqin Chen, Xuejing Liu, Jialin Wang, Wenbin Ge, Sibo Song, Kai Dang, Peng Wang, Shijie Wang, Jun Tang, Humen Zhong, Yuanzhi Zhu, Mingkun Yang, Zhaohai Li, Jianqiang Wan, Pengfei Wang, Wei Ding, Zheren Fu, Yiheng Xu, Jiabo Ye, Xi~Zhang, Tianbao Xie, Zesen Cheng, Hang Zhang, Zhibo Yang, Haiyang Xu, and Junyang Lin.
\newblock Qwen2.5-vl technical report, 2025.
\newblock URL \url{https://arxiv.org/abs/2502.13923}.

\bibitem[Baker et~al.(2017)Baker, Jara-Ettinger, Saxe, and Tenenbaum]{Baker2017RationalQA}
Chris~L. Baker, Julian Jara-Ettinger, Rebecca Saxe, and Joshua~B. Tenenbaum.
\newblock Rational quantitative attribution of beliefs, desires and percepts in human mentalizing.
\newblock \emph{Nature Human Behaviour}, 1, 2017.
\newblock URL \url{https://api.semanticscholar.org/CorpusID:3338320}.

\bibitem[Buch et~al.(2025)Buch, Nagrani, Arnab, and Schmid]{Buch2025FlexibleFS}
S.~Buch, Arsha Nagrani, Anurag Arnab, and Cordelia Schmid.
\newblock Flexible frame selection for efficient video reasoning.
\newblock \emph{2025 IEEE/CVF Conference on Computer Vision and Pattern Recognition (CVPR)}, pp.\  29071--29082, 2025.
\newblock URL \url{https://api.semanticscholar.org/CorpusID:280654792}.

\bibitem[Caba~Heilbron et~al.(2015)Caba~Heilbron, Escorcia, Ghanem, and Carlos~Niebles]{caba2015activitynet}
Fabian Caba~Heilbron, Victor Escorcia, Bernard Ghanem, and Juan Carlos~Niebles.
\newblock Activitynet: A large-scale video benchmark for human activity understanding.
\newblock In \emph{Proceedings of the IEEE/CVF Conference on Computer Vision and Pattern Recognition (CVPR)}, 2015.

\bibitem[Cheng et~al.(2024)Cheng, Leng, Zhang, Xin, Li, Chen, Zhu, Zhang, Luo, Zhao, et~al.]{cheng2024videollama2}
Zesen Cheng, Sicong Leng, Hang Zhang, Yifei Xin, Xin Li, Guanzheng Chen, Yongxin Zhu, Wenqi Zhang, Ziyang Luo, Deli Zhao, et~al.
\newblock Videollama 2: Advancing spatial-temporal modeling and audio understanding in video-llms.
\newblock \emph{arXiv preprint arXiv:2406.07476}, 2024.

\bibitem[Chinchure et~al.(2025)Chinchure, Ravi, Ng, Shwartz, Li, and Sigal]{chinchure2025black}
Aditya Chinchure, Sahithya Ravi, Raymond Ng, Vered Shwartz, Boyang Li, and Leonid Sigal.
\newblock Black swan: Abductive and defeasible video reasoning in unpredictable events.
\newblock In \emph{Proceedings of the Computer Vision and Pattern Recognition Conference}, pp.\  24201--24210, 2025.

\bibitem[Dayoon~Ko(2023)]{ko2023can}
Gunhee~Kim Dayoon~Ko, Sangho~Lee.
\newblock Can language models laugh at youtube short-form videos?
\newblock In \emph{The 2023 Conference on Empirical Methods in Natural Language Processing}, 2023.

\bibitem[Duka et~al.(2022)Duka, Kukleva, and Schiele]{duka2022leveraging}
Enea Duka, Anna Kukleva, and Bernt Schiele.
\newblock Leveraging self-supervised training for unintentional action recognition.
\newblock In \emph{European Conference on Computer Vision Workshop SSLWIN (ECCVW)}. Springer, 2022.

\bibitem[Epstein et~al.(2020)Epstein, Chen, and Vondrick]{Epstein_2020_CVPR}
Dave Epstein, Boyuan Chen, and Carl Vondrick.
\newblock Oops! predicting unintentional action in video.
\newblock In \emph{The IEEE/CVF Conference on Computer Vision and Pattern Recognition (CVPR)}, June 2020.

\bibitem[Fu et~al.(2024)Fu, Dai, Luo, Li, Ren, Zhang, Wang, Zhou, Shen, Zhang, et~al.]{fu2024video}
Chaoyou Fu, Yuhan Dai, Yondong Luo, Lei Li, Shuhuai Ren, Renrui Zhang, Zihan Wang, Chenyu Zhou, Yunhang Shen, Mengdan Zhang, et~al.
\newblock Video-mme: The first-ever comprehensive evaluation benchmark of multi-modal llms in video analysis.
\newblock \emph{arXiv preprint arXiv:2405.21075}, 2024.

\bibitem[Guo et~al.(2025)Guo, Chen, Wang, He, Xu, Ye, Sun, and Xiong]{guo2025logicinframesdynamickeyframesearch}
Weiyu Guo, Ziyang Chen, Shaoguang Wang, Jianxiang He, Yijie Xu, Jinhui Ye, Ying Sun, and Hui Xiong.
\newblock Logic-in-frames: Dynamic keyframe search via visual semantic-logical verification for long video understanding, 2025.
\newblock URL \url{https://arxiv.org/abs/2503.13139}.

\bibitem[Gupta et~al.(2025)Gupta, Corona, Ge, Wang, Klein, Darrell, and Chan]{gupta-etal-2025-enough}
Ritwik Gupta, Rodolfo Corona, Jiaxin Ge, Eric Wang, Dan Klein, Trevor Darrell, and David~M. Chan.
\newblock Enough coin flips can make {LLM}s act {B}ayesian.
\newblock In Wanxiang Che, Joyce Nabende, Ekaterina Shutova, and Mohammad~Taher Pilehvar (eds.), \emph{Proceedings of the 63rd Annual Meeting of the Association for Computational Linguistics (Volume 1: Long Papers)}, pp.\  7634--7655, Vienna, Austria, July 2025. Association for Computational Linguistics.
\newblock ISBN 979-8-89176-251-0.
\newblock \doi{10.18653/v1/2025.acl-long.377}.
\newblock URL \url{https://aclanthology.org/2025.acl-long.377/}.

\bibitem[Holtzman et~al.(2020)Holtzman, Buys, Du, Forbes, and Choi]{Holtzman2020The}
Ari Holtzman, Jan Buys, Li~Du, Maxwell Forbes, and Yejin Choi.
\newblock The curious case of neural text degeneration.
\newblock In \emph{International Conference on Learning Representations}, 2020.
\newblock URL \url{https://openreview.net/forum?id=rygGQyrFvH}.

\bibitem[Hu et~al.(2025)Hu, Gao, Nie, Zhou, Tran, Neiman, Wang, Shah, Hamid, Yin, and Chilimbi]{Hu2025MLLMBV}
Kai Hu, Feng Gao, Xiaohan Nie, Peng Zhou, Son Tran, Tal Neiman, Lingyun Wang, Mubarak Shah, Raffay Hamid, Bing Yin, and Trishul~M. Chilimbi.
\newblock M-llm based video frame selection for efficient video understanding.
\newblock \emph{2025 IEEE/CVF Conference on Computer Vision and Pattern Recognition (CVPR)}, pp.\  13702--13712, 2025.
\newblock URL \url{https://api.semanticscholar.org/CorpusID:276647361}.

\bibitem[Itti \& Baldi(2005)Itti and Baldi]{NIPS2005_0172d289}
Laurent Itti and Pierre Baldi.
\newblock Bayesian surprise attracts human attention.
\newblock In Y.~Weiss, B.~Sch\"{o}lkopf, and J.~Platt (eds.), \emph{Advances in Neural Information Processing Systems}, volume~18. MIT Press, 2005.
\newblock URL \url{https://proceedings.neurips.cc/paper_files/paper/2005/file/0172d289da48c48de8c5ebf3de9f7ee1-Paper.pdf}.

\bibitem[Kim et~al.(2025)Kim, Sclar, Zhi-Xuan, Ying, Levine, Liu, Tenenbaum, and Choi]{kim2025hypothesisdriven}
Hyunwoo Kim, Melanie Sclar, Tan Zhi-Xuan, Lance Ying, Sydney Levine, Yang Liu, Joshua~B. Tenenbaum, and Yejin Choi.
\newblock Hypothesis-driven theory-of-mind reasoning for large language models.
\newblock In \emph{Second Conference on Language Modeling}, 2025.
\newblock URL \url{https://openreview.net/forum?id=yGQqTuSJPK}.

\bibitem[Li et~al.(2024)Li, Wang, He, Li, Wang, Liu, Wang, Xu, Chen, Luo, et~al.]{li2024mvbenchvideochat2}
Kunchang Li, Yali Wang, Yinan He, Yizhuo Li, Yi~Wang, Yi~Liu, Zun Wang, Jilan Xu, Guo Chen, Ping Luo, et~al.
\newblock Mvbench: A comprehensive multi-modal video understanding benchmark.
\newblock In \emph{Proceedings of the IEEE/CVF Conference on Computer Vision and Pattern Recognition}, pp.\  22195--22206, 2024.

\bibitem[Liang et~al.(2024)Liang, Li, Bai, Huang, Sun, Wang, He, Cui, Chen, and Zhang]{Liang2024KeyVideoLLMTL}
Hao Liang, Jiapeng Li, Tianyi Bai, Xijie Huang, Linzhuang Sun, Zhengren Wang, Conghui He, Bin Cui, Chong Chen, and Wentao Zhang.
\newblock Keyvideollm: Towards large-scale video keyframe selection.
\newblock \emph{ArXiv}, abs/2407.03104, 2024.
\newblock URL \url{https://api.semanticscholar.org/CorpusID:270924158}.

\bibitem[Lin et~al.(2023)Lin, Zhang, Chen, Pramanick, Gao, Wang, Yan, and Shou]{lin2023univtg}
Kevin~Qinghong Lin, Pengchuan Zhang, Joya Chen, Shraman Pramanick, Difei Gao, Alex~Jinpeng Wang, Rui Yan, and Mike~Zheng Shou.
\newblock Univtg: Towards unified video-language temporal grounding, 2023.

\bibitem[Liu et~al.(2023)Liu, Li, Li, and Lee]{liu2023improvedllava}
Haotian Liu, Chunyuan Li, Yuheng Li, and Yong~Jae Lee.
\newblock Improved baselines with visual instruction tuning, 2023.

\bibitem[Liu et~al.(2024)Liu, Li, Li, Li, Zhang, Shen, and Lee]{liu2024llavanext}
Haotian Liu, Chunyuan Li, Yuheng Li, Bo~Li, Yuanhan Zhang, Sheng Shen, and Yong~Jae Lee.
\newblock Llava-next: Improved reasoning, ocr, and world knowledge, January 2024.
\newblock URL \url{https://llava-vl.github.io/blog/2024-01-30-llava-next/}.

\bibitem[Majumdar et~al.(2024)Majumdar, Ajay, Zhang, Putta, Yenamandra, Henaff, Silwal, Mcvay, Maksymets, Arnaud, et~al.]{majumdar2024openeqa}
Arjun Majumdar, Anurag Ajay, Xiaohan Zhang, Pranav Putta, Sriram Yenamandra, Mikael Henaff, Sneha Silwal, Paul Mcvay, Oleksandr Maksymets, Sergio Arnaud, et~al.
\newblock Openeqa: Embodied question answering in the era of foundation models.
\newblock In \emph{Proceedings of the IEEE/CVF conference on computer vision and pattern recognition}, pp.\  16488--16498, 2024.

\bibitem[Mann \& Kaur(2015)Mann and Kaur]{Mann2015KeyFE}
Jaspreet~Kaur Mann and Navjot Kaur.
\newblock Key frame extraction from a video using edge change ratio.
\newblock 2015.
\newblock URL \url{https://api.semanticscholar.org/CorpusID:52062936}.

\bibitem[Millidge et~al.(2022)Millidge, Seth, and Buckley]{millidge2022predictivecodingtheoreticalexperimental}
Beren Millidge, Anil Seth, and Christopher~L Buckley.
\newblock Predictive coding: a theoretical and experimental review, 2022.
\newblock URL \url{https://arxiv.org/abs/2107.12979}.

\bibitem[Omine et~al.(2024)Omine, Akita, and Tsuruno]{omine24_laughseg}
Taisei Omine, Kenta Akita, and Reiji Tsuruno.
\newblock Robust laughter segmentation with automatic diverse data synthesis.
\newblock In \emph{Interspeech 2024}, pp.\  4748--4752, 2024.
\newblock \doi{10.21437/Interspeech.2024-1644}.

\bibitem[OpenAI(2024)]{openai2024gpt4o}
OpenAI.
\newblock {GPT-4o} system card, 2024.

\bibitem[Radford et~al.(2023)Radford, Kim, Xu, Brockman, McLeavey, and Sutskever]{whisper}
Alec Radford, Jong~Wook Kim, Tao Xu, Greg Brockman, Christine McLeavey, and Ilya Sutskever.
\newblock Robust speech recognition via large-scale weak supervision.
\newblock In \emph{International conference on machine learning}, pp.\  28492--28518. PMLR, 2023.

\bibitem[Ren et~al.(2023)Ren, Yao, Li, Sun, and Hou]{Ren2023TimeChatAT}
Shuhuai Ren, Linli Yao, Shicheng Li, Xu~Sun, and Lu~Hou.
\newblock Timechat: A time-sensitive multimodal large language model for long video understanding.
\newblock \emph{2024 IEEE/CVF Conference on Computer Vision and Pattern Recognition (CVPR)}, pp.\  14313--14323, 2023.
\newblock URL \url{https://api.semanticscholar.org/CorpusID:265608767}.

\bibitem[Rudinger et~al.(2020)Rudinger, Shwartz, Hwang, Bhagavatula, Forbes, Le~Bras, Smith, and Choi]{rudinger-etal-2020-thinking}
Rachel Rudinger, Vered Shwartz, Jena~D. Hwang, Chandra Bhagavatula, Maxwell Forbes, Ronan Le~Bras, Noah~A. Smith, and Yejin Choi.
\newblock Thinking like a skeptic: Defeasible inference in natural language.
\newblock In Trevor Cohn, Yulan He, and Yang Liu (eds.), \emph{Findings of the Association for Computational Linguistics: EMNLP 2020}, pp.\  4661--4675, Online, November 2020. Association for Computational Linguistics.
\newblock \doi{10.18653/v1/2020.findings-emnlp.418}.
\newblock URL \url{https://aclanthology.org/2020.findings-emnlp.418/}.

\bibitem[Sclar et~al.(2023)Sclar, Kumar, West, Suhr, Choi, and Tsvetkov]{sclar-etal-2023-minding}
Melanie Sclar, Sachin Kumar, Peter West, Alane Suhr, Yejin Choi, and Yulia Tsvetkov.
\newblock Minding language models' (lack of) theory of mind: A plug-and-play multi-character belief tracker.
\newblock In Anna Rogers, Jordan Boyd-Graber, and Naoaki Okazaki (eds.), \emph{Proceedings of the 61st Annual Meeting of the Association for Computational Linguistics (Volume 1: Long Papers)}, pp.\  13960--13980, Toronto, Canada, July 2023. Association for Computational Linguistics.
\newblock \doi{10.18653/v1/2023.acl-long.780}.
\newblock URL \url{https://aclanthology.org/2023.acl-long.780/}.

\bibitem[Shao et~al.(2024)Shao, Wang, Zhu, Xu, Song, Bi, Zhang, Zhang, Li, Wu, and Guo]{shao2024deepseekmathpushinglimitsmathematical}
Zhihong Shao, Peiyi Wang, Qihao Zhu, Runxin Xu, Junxiao Song, Xiao Bi, Haowei Zhang, Mingchuan Zhang, Y.~K. Li, Y.~Wu, and Daya Guo.
\newblock Deepseekmath: Pushing the limits of mathematical reasoning in open language models, 2024.
\newblock URL \url{https://arxiv.org/abs/2402.03300}.

\bibitem[Tang et~al.(2025)Tang, Qiu, Xie, Tian, Jiao, and Ye]{tang2025adaptive}
Xi~Tang, Jihao Qiu, Lingxi Xie, Yunjie Tian, Jianbin Jiao, and Qixiang Ye.
\newblock Adaptive keyframe sampling for long video understanding.
\newblock \emph{arXiv preprint arXiv:2502.21271}, 2025.

\bibitem[V \& Narayanan(2015)V and Narayanan]{CV2015KeyframeEB}
Sheena~C V and N.K. Narayanan.
\newblock Key-frame extraction by analysis of histograms of video frames using statistical methods.
\newblock \emph{Procedia Computer Science}, 70:\penalty0 36--40, 2015.
\newblock URL \url{https://api.semanticscholar.org/CorpusID:61942704}.

\bibitem[Wang et~al.(2024)Wang, Lai, Sun, and Ge]{Wang2024WeaklySG}
Haibo Wang, Chenghang Lai, Yixuan Sun, and Weifeng Ge.
\newblock Weakly supervised gaussian contrastive grounding with large multimodal models for video question answering.
\newblock \emph{Proceedings of the 32nd ACM International Conference on Multimedia}, 2024.
\newblock URL \url{https://api.semanticscholar.org/CorpusID:267060847}.

\bibitem[Wang et~al.(2025)Wang, Yu, Stengel-Eskin, Yoon, Cheng, Bertasius, and Bansal]{Wang_2025_CVPR}
Ziyang Wang, Shoubin Yu, Elias Stengel-Eskin, Jaehong Yoon, Feng Cheng, Gedas Bertasius, and Mohit Bansal.
\newblock Videotree: Adaptive tree-based video representation for llm reasoning on long videos.
\newblock In \emph{Proceedings of the Computer Vision and Pattern Recognition Conference (CVPR)}, pp.\  3272--3283, June 2025.

\bibitem[Wolf(1996)]{Wolf1996KeyFS}
Wayne~H. Wolf.
\newblock Key frame selection by motion analysis.
\newblock \emph{1996 IEEE International Conference on Acoustics, Speech, and Signal Processing Conference Proceedings}, 2:\penalty0 1228--1231 vol. 2, 1996.
\newblock URL \url{https://api.semanticscholar.org/CorpusID:7256933}.

\bibitem[Xiao et~al.(2021)Xiao, Shang, Yao, and Chua]{xiao2021next}
Junbin Xiao, Xindi Shang, Angela Yao, and Tat-Seng Chua.
\newblock Next-qa: Next phase of question-answering to explaining temporal actions.
\newblock In \emph{Proceedings of the IEEE/CVF Conference on Computer Vision and Pattern Recognition (CVPR)}, pp.\  9777--9786, June 2021.

\bibitem[Xie et~al.(2025)Xie, Zhang, Zhou, Li, Zhang, Hessel, Yang, and Liu]{xie2025funqa}
Binzhu Xie, Sicheng Zhang, Zitang Zhou, Bo~Li, Yuanhan Zhang, Jack Hessel, Jingkang Yang, and Ziwei Liu.
\newblock Funqa: Towards surprising video comprehension.
\newblock In \emph{European Conference on Computer Vision}, pp.\  39--57. Springer, 2025.

\bibitem[Yu et~al.(2025)Yu, JIN, Wang, Chen, Jin, ZUO, XIAOLEI, Sun, Zhang, Wu, Zhang, and Sun]{yu2025framevoyager}
Sicheng Yu, CHENGKAI JIN, Huanyu Wang, Zhenghao Chen, Sheng Jin, ZHONGRONG ZUO, XU~XIAOLEI, Zhenbang Sun, Bingni Zhang, Jiawei Wu, Hao Zhang, and Qianru Sun.
\newblock Frame-voyager: Learning to query frames for video large language models.
\newblock In \emph{The Thirteenth International Conference on Learning Representations}, 2025.
\newblock URL \url{https://openreview.net/forum?id=LNL7zKvm7e}.

\end{thebibliography}
\bibliographystyle{iclr2026_conference}

\clearpage
\setcounter{page}{1}
\renewcommand{\thesection}{\Alph{section}}
\setcounter{section}{0}
\appendix
\renewcommand{\thefigure}{A\arabic{figure}}
\setcounter{figure}{0}


\section{Prompts}
\label{app:allprompts}
\subsection{Hypothesis Prompts}
\label{app:hypothesis_prompt}
\textbf{Generation.} 
We prompt the model with a memory of prior events and recent frames, asking for a concise next–frame prediction (8--10 words):

\begin{prompt}
Given a textual summary of the video so far and the most recent \emph{prior window of frames}, predict what will most likely happen in the next frame. 

\textbf{Context so far:}
{memory\_text}

\textbf{Prior window (video inputs):}
\emph{A sequence of images corresponding to the last W frames.}

\textbf{Output format:}
Hypothesis: 8--10 words
\end{prompt}

\textbf{Prior.} We use the following prompt to score each hypothesis.
\label{app:scoring_prompt}
\begin{prompt}
\textbf{Context so far:}  
{memory\_text}  

\textbf{Prior window (video inputs):}  
A sequence of images corresponding to the last $W$ frames.  

\textbf{Current frame:}  
The observed frame immediately following the prior window.  

Here is what will happen next: [hypothesis statement]         
\end{prompt}

\textbf{Posterior.} We use the following prompt to score each hypothesis and compute the probability of yes as the posterior likelihood of that hypothesis.
\label{app:scoring_prompt}
\begin{prompt}
You are given a textual summary of the video so far, a \emph{prior window} of frames, and the \emph{current frame} that follows.  
Your task is to evaluate whether each hypothesis generated from the prior context still holds in the current frame.  

\textbf{Context so far:}  
{memory\_text}  

\textbf{Prior window (video inputs):}  
A sequence of images corresponding to the last $W$ frames.  

\textbf{Current frame:}  
The observed frame immediately following the prior window.  

\textbf{Hypothesis:} [hypothesis statement]  

Question: Is this hypothesis true in the \emph{current frame}?  
Answer with a single word: \texttt{yes} or \texttt{no}.         
\end{prompt}

\subsection{LLM Reward Prompt}
\label{app:llm_reward}

\begin{prompt}
Rate how closely the content of the prediction matches the content of the reference description in terms of meaning and how well it captures important details regarding events in the video.
Ignore the difference in length.
Score 0.0-1.0 where:

0.0-0.3: Poor match (key details in the reference are missing in the prediction)
0.4-0.6: Moderate match (a few key details in the reference are captured in the prediction)
0.7-0.9: Good match (most key details are present in the prediction)
1.0: Perfect match (all key details in the reference are accurately captured in the prediction)
Output only the numerical score (e.g., 0.75).

\textbf{Reference}: {gt}\\
\textbf{Response}: {response}

Score:
\end{prompt}
\subsection{Zero-shot scorer prompt}
\label{app:qwen2.5_surprise_prompt}
\begin{prompt}
You are analyzing video frames for surprisingness. For each frame, assign a label of 1 if it is surprising and 0 if it is not.\\
1: surprising content\\
0: expected content\\

\textbf{Video frames}: \emph{Original Video Frames}
\end{prompt}

\section{Historical Summmary}
\label{app:historical_summary}
In our implementation, the memory of what happened since the beginning of the video i.e the Historical summary,  is maintained as a rolling textual summary that updates with each newly observed frame. Before use, the memory is compressed using the BART-Large-CNN summarization model whenever it exceeds approximately 200 word. For each step, the model receives the condensed memory, a short window of prior frames, and the most recent observed frame, and generates a caption describing the new event. This caption is appended to the memory, yielding a continuously updated narrative of ``what has happened so far'', which is then used for hypothesis generation and surprise computation.

\section{Hyperparameters}
\label{app:Hyperparams}
\subsection{Training}
\label{app:Hyperparams:train}
We train using 4 H100s on a single node with DeepSpeed ZeRO-3 offload. All runs use Qwen2.5-VL-7B-Instruct as the backbone, with FlashAttention-2, bfloat16 precision, and PEFT enabled.   

\begin{table}[h]
\centering
\caption{Key hyperparameters for GRPO training.}
\label{tab:train_hparams}
\small
\setlength{\tabcolsep}{8pt}
\renewcommand{\arraystretch}{1.2}
\begin{tabular}{ll}
\toprule
Hyperparameter & Value \\
\midrule
Learning rate                & $1 \times 10^{-6}$ \\
GRPO $\beta$                 & 0.1 \\
Number of GRPO Rollouts  & 3 \\
Number of Hypotheses per window  & 3 \\
Max prompt length            & 8192 tokens \\
Training samples             & 2000 \\
Epochs                       & 1 \\
Per-device batch size        & 1 \\
Effective global batch size  & 4 \\
Random seed                  & 42 \\

\bottomrule
\end{tabular}
\end{table}
\subsection{Inference}
\label{app:Hyperparams:inference}
For both \scorer{} and \rlalgo{}, we maintain a hypothesis set $N = 3$ per time step. We use a prior window of $W=4$ frames, and the frames for surprise scoring are allocated in proportion to the video duration,  $F = f(\text{duration})$. Videos up to a minute are assigned a base budget of 8 frames. For longer videos, the budget continues to double with each additional minute.

\section{Surprise Localization Metrics}
\label{app:metrics_surprise}
\textbf{Accuracy@$\delta$.}
Let $\hat{t}$ be the predicted time (in seconds) obtained by converting the frame with the highest surprise score to time, and let $t^\star$ be the ground-truth transition time. We use the transition time provided in Oops! directly. For FunQA and Mr.Bean, center of the most surprising window is used as transition time. The instance-level score is
\[
\text{Accuracy@}\delta \;=\; \mathbb{1}\!\left[\,|\hat{t} - t^\star| \le \delta\,\right],
\]
and the reported metric is the mean of this indicator over the evaluation videos. Typical choices include $\delta\!\in\!\{0.25,1.0\}$ seconds.

\textbf{IoU.}
Let $\mathcal{W}_{\text{pred}} = \{[a,b] : s(t) > \tau \text{ for } t \in [a,b]\}$ be the predicted surprising windows and $\mathcal{W}_{\text{gt}}$ be the given set of ground truth surprising windows. The Temporal IoU is:
\[
\text{Temporal IoU} = \frac{\text{intersection coverage}}{\text{union coverage}} = \frac{|\bigcup \mathcal{W}_{\text{pred}} \cap \bigcup \mathcal{W}_{\text{gt}}|}{|\bigcup \mathcal{W}_{\text{pred}} \cup \bigcup \mathcal{W}_{\text{gt}}|}
\]
where $|\cdot|$ denotes temporal coverage (total duration). We define predicted surprising windows as a set of maximal contiguous intervals where the surprise score exceeds a threshold $\tau = 0.8 \times \max_t s(t)$ 
for that video.

\section{Mr.\ Bean}
\label{sec:app:bean}

We collect 48 videos from Mr. Bean compilation videos on YouTube. Specifically, we follow this process:
\begin{enumerate}
    \item Each clip is divided into its scenes using a scene detector model, PySceneDetect, using its ContentDetector\footnote{\url{https://www.scenedetect.com/docs/0.6.1/api/detectors.html}}, with a threshold of 30. 
    \item Scenes shorter than 12 seconds and longer than 60 seconds are filtered out, to reduce incorrect scene cuts or have videos that are too short for our analysis. 
    \item We extract the audio from these scenes, and use a laughter segmentation model from \citet{omine24_laughseg} to identify where laughter is present. We filter scenes to obtain only those that have 1 to 3 laughter segments.
    \item Because we rely on laughter tracks as our silver-standard surprise annotation, we transcribe the audio in these clips. We use OpenAI's Whisper \citep{whisper}, with the \textit{turbo} model. If a clip has too many words in its transcription ($>8$), it is discarded. Through empirical observation, we found that laughter occurs in small peaks. We ensure that at least one such loud peak ($>-28dB$) of at least 1 second occurs.
    \item As a final step, we manually filter through the video set to discard scenes which contain additional noises (e.g. bells) or scenes that are not semantically meaningful (e.g. the opening credits) that may have passed the other filters. This leaves us with 48 video clips. 
\end{enumerate}

The full list of clips, a link to their original source, along with video scenes which we use, will be provided with the code and data release.

\section{Complexity Analysis}
\label{sec:complexity}
Let a video contain $T$ frames. We uniformly sample a fixed budget of $F$ frames, so the video is divided into $W = T/F$ segments and one frame is drawn from each segment. 
For each sampled frame we generate $N$ text hypotheses and compute their prior and posterior likelihoods.

\paragraph{Time Complexity.}
The method requires $F$ hypothesis-generation steps and two batched likelihood evaluations per step. 
The total cost is therefore
\[
O(F \cdot N),
\]
which is linear in the chosen frame budget $F$ (and therefore at most linear in $T$ if $F$ grows with $T$).
\paragraph{Relation to Inference-Time Scaling.}
Our overhead is comparable to recent inference-time scaling methods for Video-LLMs: a controllable number of extra forward passes improves where the model allocates its fixed frame budget, without changing its architecture.
\paragraph{Interpretability.}
Because \scorer\ represents beliefs as \emph{textual hypotheses}, its Bayesian surprise scores are interpretable: one can inspect the generated hypotheses to understand what the model ``expected'' versus what the new frames revealed.

\section{JSD}
\label{app:jsd}
For bounded and symmetric reporting, we convert KL to the Jensen–Shannon divergence (JSD), where,
\begin{align}
\label{eq:bayesian-surprise-JSD}
    \mathcal{S}_t \;&= \mathrm{JSD}(P_{\text{post}},P_{\text{prior}})=\tfrac{1}{2}D_{\mathrm{KL}}(P_{\text{post}} \!\|\! M)+\tfrac{1}{2}D_{\mathrm{KL}}(P_{\text{prior}} \!\|\! M),
\end{align}

where $M=\tfrac{1}{2}(P_{\text{post}}+P_{\text{prior}})$, which maps naturally to $[0,1]$ after $\log_2$ normalization.

\section{Human Evaluation Template}
\label{app:human_eval_temp}
\begin{figure}[t!]
    \centering
    \includegraphics[width=1\linewidth]{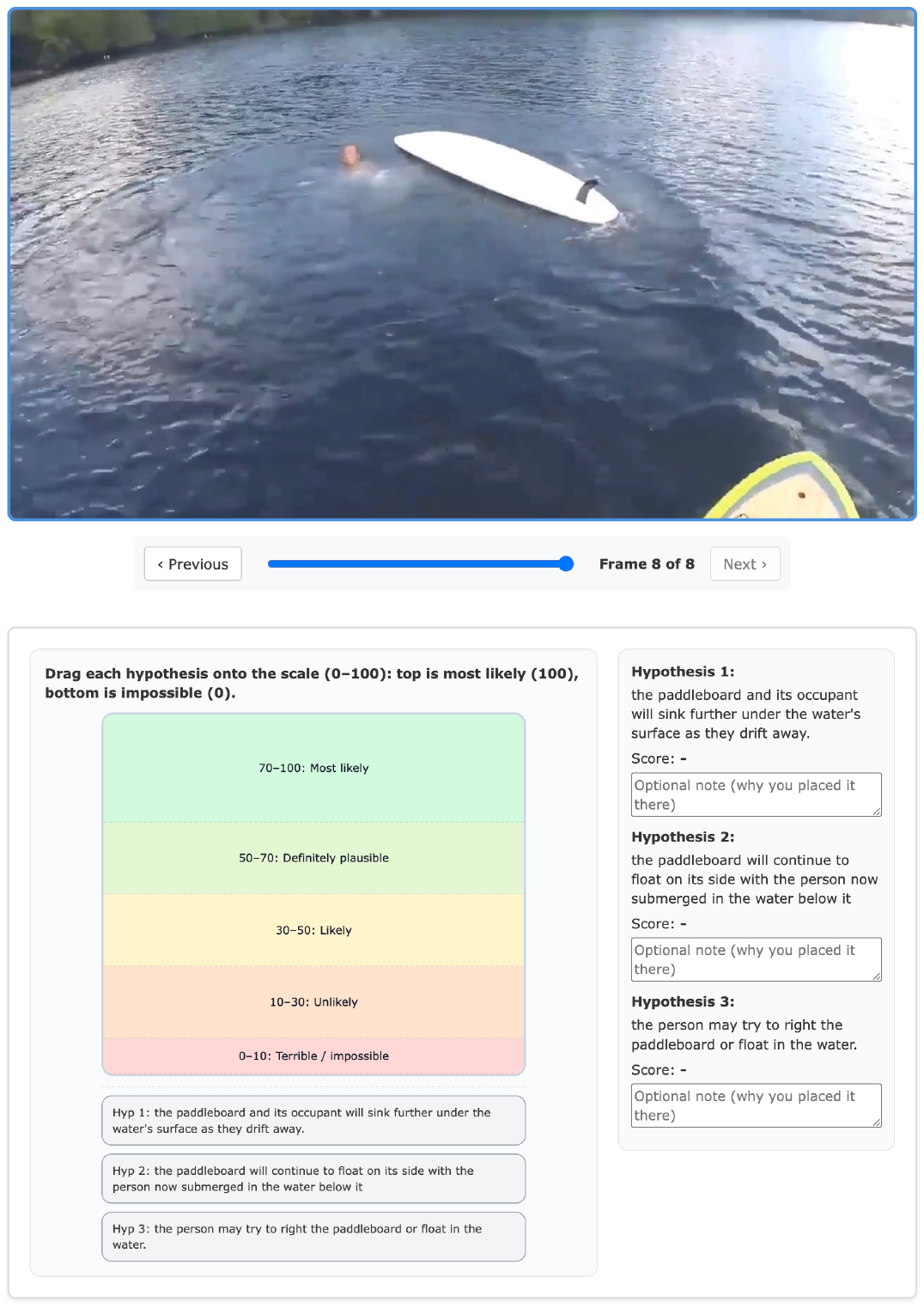}
    \caption{We ask human evaluators to score the hypotheses by dragging and dropping them into likelihood bands between 0 -- 100. This is repeated twice -- by scoring the hypothesis with and without the observed new frame.}
    \label{fig:humaneval}
\end{figure}
\begin{figure}[b!]
    \centering
    \includegraphics[width=\linewidth]{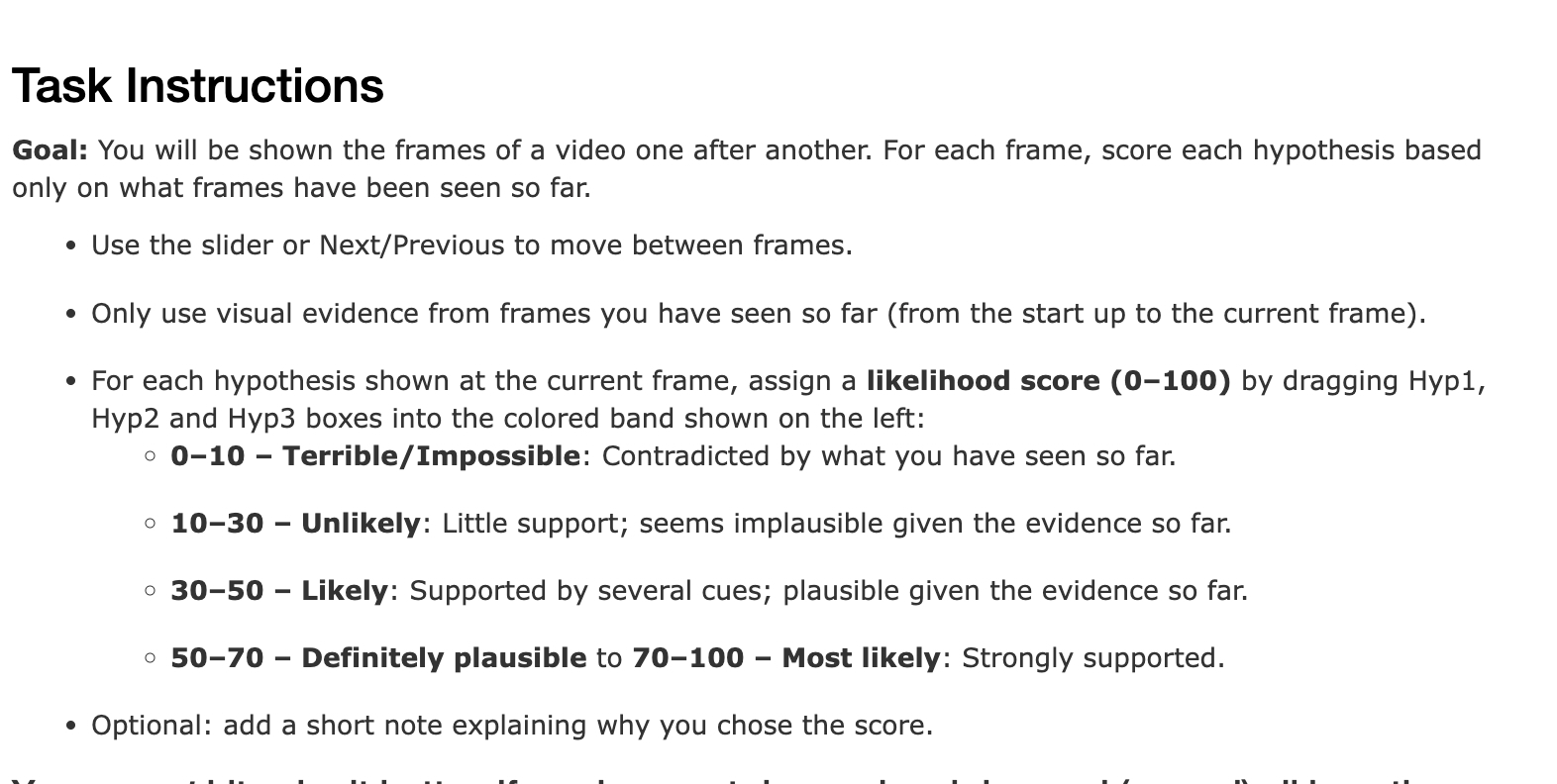}
    \caption{Instructions shown to human evaluators.}
    \label{fig:ins}
\end{figure}

Fig ~\ref{fig:humaneval} and Fig~\ref{fig:ins} show the template and instructions used for human evaluation.

\end{document}